\documentclass{SCIS2019}
\usepackage{lineno,hyperref}
\usepackage{latexsym}
\usepackage{mathtools}
\usepackage{mathrsfs}
\usepackage{amsthm}
\usepackage{amsfonts}
\usepackage{amssymb}
\usepackage{amsmath}
\usepackage{enumerate}
\usepackage{extarrows}
\usepackage{xcolor}
\usepackage{cases}
\usepackage{array}
\usepackage{url}
\usepackage{dsfont}
\usepackage{threeparttable}
\usepackage{multirow}
\usepackage{booktabs}
\usepackage{epstopdf}

\def\Tr{\mathop{\rm Tr}}
\def\vec{\mathop{\rm vec}}

\def\bcirc{\mathop{\rm bcirc}}
\def\rank{\mathop{\rm rank}}
\def\fold{\mathop{\rm fold}}
\def\unfold{\mathop{\rm unfold}}
\begin{document}
\ArticleType{RESEARCH PAPER}
\Year{2019}
\Month{}
\Vol{}
\No{}
\DOI{}
\ArtNo{}
\ReceiveDate{}
\ReviseDate{}
\AcceptDate{}
\OnlineDate{}

\title{Tensor Restricted Isometry Property Analysis For a Large Class of Random Measurement Ensembles}{Tensor Restricted Isometry Property Analysis For a Large Class of Random Measurement Ensembles}

\author[1]{Feng ZHANG}{}
\author[1]{Wendong WANG}{}
\author[1]{Jingyao HOU}{}
\author[2]{Jianjun WANG}{{wjj@swu.edu.cn}}
\author[1]{Jianwen HUANG}{}

\AuthorMark{Feng ZHANG}

\AuthorCitation{Feng ZHANG, Wendong WANG, Jingyao HOU, et al}


\address[1]{School of Mathematics and Statistics, Southwest University, Chongqing {\rm 400715}, China}
\address[2]{College of Artificial Intelligence, Southwest University, Chongqing {\rm 400715}, China}

\abstract{In previous work, theoretical analysis based on the tensor Restricted Isometry Property (t-RIP) established the robust recovery guarantees of a low-tubal-rank tensor. The obtained sufficient conditions depend strongly on the assumption that the linear measurement maps satisfy the t-RIP. In this paper, by exploiting the probabilistic arguments, we prove that such linear measurement maps exist under suitable conditions on the number of measurements in terms of the tubal rank $r$ and the size of third-order tensor $n_{1}$, $n_{2}$, $n_{3}$. The obtained minimal possible number of linear measurements is very reasonable and order optimal from the perspective of the degrees of freedom of a tensor with tubal rank $r$. Especially, we consider a random sub-Gaussian distribution that includes zero-mean Gaussian distributions, symmetric Bernoulli distributions and all zero-mean bounded distributions and construct a large class of linear maps that satisfy a t-RIP with high probability. Moreover, the validity of the required number of measurements is verified by numerical experiments.}


\maketitle

\section{Introduction}\label{Introduction}
Low-Rank Tensor Recovery (LRTR) \cite{Lu2018Exact,rauhut2017low,goldfarb2014robust}, as a natural higher-order generalization of Compressed Sensing (CS) \cite{donoho2006compressed,Candes2005Decoding,wang2019nonconvex,wang2016coherence} and Low-Rank Matrix Recovery (LRMR) \cite{candes2011tight,nie2012low,recht2010guaranteed}, is being extensively applied in various fields of artificial intelligence, including computer vision \cite{xie2018guaranteed}, image processing \cite{Jiang2019Fastderain} and machine learning \cite{Peng2014Decomposable}, etc. LRTR aims at recovering a low-rank tensor $\boldsymbol{\mathcal{X}}\in\mathbb{R}^{n_{1} \times n_{2} \times n_{3}}$ (this work focuses on the third-order tensor) from linear noise measurements $\boldsymbol{y}=\boldsymbol{\mathfrak{M}}(\boldsymbol{\mathcal{X}})+\boldsymbol{w}$, where $\boldsymbol{\mathfrak{M}}$ is a random map from $\mathbb{R}^{n_{1} \times n_{2} \times n_{3}}$ to $\mathbb{R}^{m}$ $(m\ll n_{1}n_{2}n_{3})$ and $\boldsymbol{w}\in\mathbb{R}^{m}$ is a vector of measurement errors with noise level $\|\boldsymbol{w}\|_{2}\leq\epsilon$.

It's not easy to achieve this goal. On one hand, the naive approach of solving the nonconvex program
\begin{equation}\label{rank min}
    \min \limits_{{\boldsymbol{\mathcal{X}}}\in\mathbb{R}^{n_{1} \times n_{2} \times n_{3}}}~\rank(\boldsymbol{\mathcal{X}}),~~s.t.~~\|\boldsymbol{y}-\boldsymbol{\mathfrak{M}}(\boldsymbol{\mathcal{X}})\|_{2}\leq\epsilon
\end{equation}
is NP-hard in general, where the operation $\rank(\boldsymbol{\mathcal{X}})$ acts as a sparsity regularization
of tensor singular values of $\boldsymbol{\mathcal{X}}$. On the other hand, some existing tensor ranks do not work well, such as CP rank \cite{kiers2000towards} and Tucker rank \cite{Tucker1966Some}. The reason for this is that the calculation of CP rank of a tensor is usually NP-hard \cite{hillar2013most} and the convex surrogate of the Tucker rank, Sum of Nuclear Norms (SNN) \cite{liu2013tensor}, is not the tightest convex relaxation. To avoid these defects, Lu et al. \cite{lu2018tensor} first pay attention to the novel tensor tubal rank of $\boldsymbol{\mathcal{X}}$ (see Definition \ref{Tensor tubal rank}), denoted as $\rank_{t}(\boldsymbol{\mathcal{X}})$, induced by tensor-tensor product (t-product) \cite{kilmer2013third} and tensor Singular Value Decomposition (t-SVD) \cite{Kilmer2011Factorization} and have proved that most of the original tensor data, including color image data, video data and face data, etc. have a low-tubal-rank structure. So some related problems such as image denoising, video foreground and background segmentation, face recognition and so on can be solved effectively by t-SVD and low-tubal-rank methods. Further, Lu et al. consider the following convex Tensor Nuclear Norm Minimization (TNNM) model
\begin{equation}\label{tensor nuclear norm min}
  \min \limits_{{\boldsymbol{\mathcal{X}}}\in\mathbb{R}^{n_{1} \times n_{2} \times n_{3}}}~\|\boldsymbol{\mathcal{X}}\|_{\circledast},~~s.t.~~\|\boldsymbol{y}-\boldsymbol{\mathfrak{M}}(\boldsymbol{\mathcal{X}})\|_{2}\leq\epsilon,
\end{equation}
where $\|\boldsymbol{\mathcal{X}}\|_{\circledast}$ is referred to as Tensor Nuclear Norm (TNN) (see Definition \ref{Tensor nuclear norm}) which has been proved to be the convex envelop of tensor average rank\footnote{The reference \cite{lu2018tensor} indicates that low average rank assumption is a weaker low tubal rank assumption, i.e., a tensor with low tubal always has a low average rank. Its definition can be found in \cite{lu2018tensor}.} within the unit ball of the tensor spectral norm \cite{lu2018tensor}. In order to facilitate the design of algorithms and the needs of practical applications, in previous work \cite{Zhang2019RIP}, Zhang et al. first present a theoretical analysis for Regularized Tensor Nuclear Norm Minimization (RTNNM) model, which takes the form
\begin{equation}\label{tensor nuclear norm min unconstrained}
  \min \limits_{{\boldsymbol{\mathcal{X}}}\in\mathbb{R}^{n_{1} \times n_{2} \times n_{3}}}~\|\boldsymbol{\mathcal{X}}\|_{\circledast}+\frac{1}{2\lambda}\|\boldsymbol{y}-\boldsymbol{\mathfrak{M}}(\boldsymbol{\mathcal{X}})\|_{2}^{2}
\end{equation}
with a positive parameter $\lambda$. Especially, the RTNNM model (\ref{tensor nuclear norm min unconstrained}) is more applicable than the constrained-TNNM model (\ref{tensor nuclear norm min}) when the noise level is not given or cannot be accurately estimated. The tensor Restricted Isometry Property (t-RIP) was first defined based on t-SVD in \cite{Zhang2019RIP} as an analysis framework for LRTR via (\ref{tensor nuclear norm min unconstrained}). For an integer $r$, the $r$-tensor restricted isometry constants of a linear map $\boldsymbol{\mathfrak{M}}: \mathbb{R}^{n_{1} \times n_{2} \times n_{3}}\rightarrow \mathbb{R}^{m}$ is defined as the smallest constants satisfying
\begin{equation}\label{t-RIP}
(1-\delta_{r})\|\boldsymbol{\mathcal{X}}\|_{F}^{2}\leq\|\boldsymbol{\mathfrak{M}}(\boldsymbol{\mathcal{X}})\|_{2}^{2}\leq(1+\delta_{r})\|\boldsymbol{\mathcal{X}}\|_{F}^{2}
\end{equation}
for all tensors $\boldsymbol{\mathcal{X}}\in\mathbb{R}^{n_{1} \times n_{2} \times n_{3}}$ whose tubal rank is at most $r$. Moreover, our Theorem 4.1 in \cite{Zhang2019RIP} shows that if $\boldsymbol{\mathfrak{M}}$ satisfies the t-RIP with $\delta_{tr}<\sqrt{(t-1)/(n_{3}^{2}+t-1)}$ for certain $t>1$, the solution to (\ref{tensor nuclear norm min unconstrained}) can robustly recover the low-tubal-rank tensor $\boldsymbol{\mathcal{X}}$.

Note that Zhang et al. \cite{Zhang2019RIP} have derived a deterministic condition of robust recovery for the RTNNM model (\ref{tensor nuclear norm min unconstrained}) based on the t-RIP. Unfortunately, it is unknown how to construct a linear map $\boldsymbol{\mathfrak{M}}$ that satisfies t-RIP. The purpose of this paper is precisely to show their existence under suitable conditions on the number of measurements in terms of the tubal rank $r$ and the size of tensor $n_{1}, n_{2} , n_{3}$ using probabilistic arguments. We consider the sub-Gaussian measurement ensemble whose all elements (tensors with size $n_{1} \times n_{2} \times n_{3} \times m$) are drawn independently according to a sub-Gaussian distribution. This includes zero-mean Gaussian distributions, symmetric Bernoulli distributions and all zero-mean bounded distributions. For such linear maps, the t-RIP holds with high probability in the stated parameter regime.

In 2018, Lu et al. \cite{Lu2018Exact} provided an exact recovery result based on the Gaussian width for TNNM model \eqref{tensor nuclear norm min}. Specifically, they pointed out that the unknown tensor of size $n_{1} \times n_{2} \times n_{3}$ with tubal rank $r$ can be exactly recovered with high probability by solving \eqref{tensor nuclear norm min} when the given number of Gaussian measurements is of the order $O(r(n_{1} + n_{2} - r)n_{3})$. In 2019, Wang et al. \cite{wang2019generalized} presented a generalized tensor Dantzig selector for low-tubal-rank tensor recovery problem with noisy measurements $\boldsymbol{y}=\boldsymbol{\mathfrak{M}}(\boldsymbol{\mathcal{X}})+\boldsymbol{w}$ where $\boldsymbol{w}$ is the noise term. They showed that when the sample size $m=\Omega(r(n_{1} + n_{2} - r)n_{3})$, the solution $\boldsymbol{\hat{\mathcal{X}}}$ of generalized tensor Dantzig selector satisfies $\|\boldsymbol{\hat{\mathcal{X}}}-\boldsymbol{\mathcal{X}}\|_{F}^{2}\leq O(r(n_{1} + n_{2} - r)n_{3}m^{-1})$ with high probability. In the noiseless setting (i.e., $\boldsymbol{w}=\boldsymbol{0}$), their results will degenerate to Lu's case. All recovery results mentioned are probabilistic. Some deterministic results involved tensor RIP have emerged in LRTR. In 2013, the first tensor deterministic condition---tensor RIP based on Tucker decomposition \cite{Tucker1966Some} which can guarantee that a given linear map $\boldsymbol{\mathfrak{M}}$ can be utilized for LRTR was proposed by Shi et al. \cite{shi2013guarantees}. They showed that a tensor $\boldsymbol{\mathcal{X}}\in\mathbb{R}^{n_{1} \times n_{2} \times n_{3}}$ with Tucker rank-$(r_{1}, r_{2}, r_{3})$ can be exactly recovered in the noiseless case if the linear map $\boldsymbol{\mathfrak{M}}$ satisfies the tensor RIP with the constant $\delta_{\Lambda}<0.4931$ for $\Lambda\in\left\{(2r_{1}, n_{2}, n_{3}),(n_{1},2r_{2}, n_{3}),(n_{1}, n_{2}, 2r_{3})\right\}$. Such tensor RIP is hardly practical because it depends on a rank tuple that differs greatly from the definition of familiar matrix rank, which will result in that some existing analysis tools and techniques can not be used for tensor cases. What's more, which linear mappings satisfy such tensor RIP is still an open problem for them.

In previous work \cite{Zhang2019RIP}, Zhang et al. used the t-RIP to answer under what conditions the robust solution to model \eqref{tensor nuclear norm min unconstrained} can be obtained. In this paper, we continue the work and answer a quintessential and all-important question: which linear maps $\boldsymbol{\mathfrak{M}}$ satisfy the t-RIP? The practical significance of this topic is to provide theoretical support for the robust recovery of low-tubal-rank tensor data from its as few as measurements in some real problems, such as Magnetic Resonance Imaging (MRI) \cite{feng2016compressive}, hyper-spectral imaging \cite{wang2017compressive}, and video security monitoring \cite{cao2016total}, etc. We take MRI as an example to illustrate. Imaging speed is important in many MRI applications. However, the speed depends largely on the amount of data collected in MRI. If the reconstructed image with high resolution can be obtained by using a small amount of data, then we can reduce the scanning time, sampling cost and pain of patients. So how to design the sampling operator and how many samples to ensure the accurate estimation of the image on the hardware side are all problems that need to be solved. Thus our research results will provide some theoretical guarantees to similar application environments. Our main contributions are summarized as follows:

Using the arguments of covering numbers and chaos processes as well as concentration inequalities, we consider a large class of sub-Gaussian distributions that include zero-mean Gaussian distributions, symmetric Bernoulli distributions and all zero-mean bounded distributions. Especially, we show that the sampling number $O(r(n_{1}+n_{2}+1)n_{3})$ is sufficient for a random sub-Gaussian measurement ensemble $\boldsymbol{\mathfrak{M}}: \mathbb{R}^{n_{1} \times n_{2} \times n_{3}}\rightarrow \mathbb{R}^{m}$ that satisfies a t-RIP at tubal rank $r$ with high probability. Moreover, the bound is reasonable and order optimal when compared with the degrees of freedom $r(n_{1}+n_{2}-r)n_{3}$ of the tubal rank-$r$ tensor of size $n_{1} \times n_{2} \times n_{3}$. In order to verify our conclusions, some numerical experiments are performed to study the variation of success recovery ratio and relative error in term of increasing measurements.

The remainder of the paper is organized as follows. In Section \ref{Notations and Preliminaries}, we introduce some notations and definitions. In Section \ref{Probabilistic tools}, some probabilistic tools for proving are given. In Section \ref{Main results}, our main results and their proofs are presented and discussed. Section \ref{Numerical experiments} conducts some numerical experiments to support our analysis. The conclusion is addressed in Section \ref{Conclusion}.

\section{Notations and preliminaries}
\label{Notations and Preliminaries}

For the sake of brevity, we list main notations which will be used later in Table \ref{notations}. For a third-order tensor $\boldsymbol{\mathcal{X}}\in\mathbb{R}^{n_{1}\times n_{2}\times n_{3}}$, let $\boldsymbol{\bar{\mathcal{X}}}$ be the Discrete Fourier transform (DFT) along the third dimension of $\boldsymbol{\mathcal{X}}$, i.e., $\boldsymbol{\bar{\mathcal{X}}}=\rm fft(\boldsymbol{\mathcal{X}},[\;],3)$. Utilizing the inverse DFT, $\boldsymbol{\mathcal{X}}$ can be calculated from $\boldsymbol{\bar{\mathcal{X}}}$ by $\boldsymbol{\mathcal{X}}=\rm ifft(\boldsymbol{\bar{\mathcal{X}}},[\;],3)$. Let $\boldsymbol{\bar{X}}\in\mathbb{R}^{n_{1}n_{3}\times n_{2}n_{3}}$ be the block diagonal matrix with each block on diagonal as the frontal slice $\boldsymbol{\bar{X}}^{(i)}$ of $\boldsymbol{\bar{\mathcal{X}}}$ and $\bcirc(\boldsymbol{\mathcal{X}})\in\mathbb{R}^{n_{1}n_{3}\times n_{2}n_{3}}$ be the block circular matrix, i.e.,
\begin{equation*}
\boldsymbol{\bar{X}}=\text{bdiag}(\boldsymbol{\bar{\mathcal{X}}})=\left(
                                                                    \begin{array}{cccc}
                                                                      \boldsymbol{\bar{X}}^{(1)} &  &  &  \\
                                                                       & \boldsymbol{\bar{X}}^{(2)} &  &  \\
                                                                       &  & \ddots &  \\
                                                                       &  &  & \boldsymbol{\bar{X}}^{(n_{3})} \\
                                                                    \end{array}
                                                                  \right)\;~\rm and~\;
\bcirc(\boldsymbol{\mathcal{X}})=\left(
                                                                    \begin{array}{cccc}
                                                                      \boldsymbol{X}^{(1)} & \boldsymbol{X}^{(n_{3})} & \cdots & \boldsymbol{X}^{(2)} \\
                                                                      \boldsymbol{X}^{(2)} & \boldsymbol{X}^{(1)} & \cdots & \boldsymbol{X}^{(3)} \\
                                                                      \vdots & \vdots & \ddots & \vdots \\
                                                                      \boldsymbol{X}^{(n_{3})} & \boldsymbol{X}^{(n_{3}-1)} & \cdots & \boldsymbol{X}^{(1)} \\
                                                                    \end{array}
                                                                  \right).
\end{equation*}
The $\rm unfold$ operator and its inverse operator $\rm fold$ are, respectively, defined as
\begin{equation*}
\unfold(\boldsymbol{\mathcal{X}})=\left(
  \begin{array}{cccc}
    \boldsymbol{X}^{(1)} \\ \boldsymbol{X}^{(2)} \\ \vdots \\ \boldsymbol{X}^{(n_{3})} \\
  \end{array}
\right)\;\;\textrm{and}\;\;
\fold(\unfold(\boldsymbol{\mathcal{X}}))=\boldsymbol{\mathcal{X}}.
\end{equation*}
The tensor transpose \cite{Kilmer2011Factorization} of $\boldsymbol{\mathcal{X}}\in\mathbb{R}^{n_{1}\times n_{2}\times n_{3}}$, denoted as $\boldsymbol{\mathcal{X}}^{T}\in\mathbb{R}^{n_{2}\times n_{1}\times n_{3}}$, is obtained by transposing each of the frontal slice and then reversing the order of transposed frontal slices 2 through $n_{3}$. The identity tensor \cite{Kilmer2011Factorization} $\boldsymbol{\mathcal{I}}\in\mathbb{R}^{n\times n\times n_{3}}$ is the tensor whose first frontal slice is the $n\times n$ identity matrix, and other frontal slices are all zeros. For tensors $\boldsymbol{\mathcal{A}}\in\mathbb{R}^{n_{1}\times n_{2}\times n_{3}}$ and $\boldsymbol{\mathcal{B}}\in\mathbb{R}^{n_{2}\times n_{4}\times n_{3}}$, the tensor-tensor product (t-product) \cite{Kilmer2011Factorization}, $\boldsymbol{\mathcal{A}}\ast\boldsymbol{\mathcal{B}}=\fold(\bcirc(\boldsymbol{\mathcal{A}})\cdot\unfold(\boldsymbol{\mathcal{B}}))$, is defined to be a tensor of size $n_{1}\times n_{4}\times n_{3}$. The orthogonal tensor \cite{Kilmer2011Factorization} $\boldsymbol{\mathcal{Q}}\in\mathbb{R}^{n\times n\times n_{3}}$ is the tensor which satisfies $\boldsymbol{\mathcal{Q}}^{T}\ast\boldsymbol{\mathcal{Q}}=\boldsymbol{\mathcal{Q}}\ast\boldsymbol{\mathcal{Q}}^{T}=\boldsymbol{\mathcal{I}}$.
A tensor is called F-diagonal \cite{Kilmer2011Factorization} if each of its frontal slices is a diagonal matrix.

\renewcommand\arraystretch{1.6}
\begin{table*}[htb] 
\centering
\fontsize{7}{7}\selectfont
\caption{Summary of main notations in the paper.} 
\begin{tabular}{p{1cm}<{\centering}|p{1.5cm}<{\centering}|p{2.2cm}<{\centering}|p{3.8cm}<{\centering}|p{1.1cm}<{\centering}|p{3.8cm}<{\centering}} 
\hline
\hline
Format & Description & Format & Description & Format & Description \\ 
\hline 
$\boldsymbol{\mathcal{X}}$ & A tensor. & $\tilde{\mathbb{W}}$ & A subset of $\mathbb{W}$. & $\boldsymbol{\mathcal{X}}(:,j,:)$ & The $j$-th lateral slice of $\boldsymbol{\mathcal{X}}$\\
\hline
$\boldsymbol{X}$ & A matrix. & $\tilde{\mathbb{W}}_{\varepsilon}$ & An $\varepsilon$-net of $\mathbb{W}$. & $\boldsymbol{\mathcal{X}}(i,j,:)$ & The tube fiber of $\boldsymbol{\mathcal{X}}$.\\
\hline
$\boldsymbol{x}$ & A vector. & $\boldsymbol{\mathcal{I}}$ & The identity tensor. & $\boldsymbol{\mathcal{X}}^{T}$ & The transpose of $\boldsymbol{\mathcal{X}}$.\\
\hline
$x$ & A scalar. & $\boldsymbol{\mathcal{X}}(i,j,k)$ or $\boldsymbol{\mathcal{X}}_{ijk}$ & The $(i, j, k)$-th entry of $\boldsymbol{\mathcal{X}}$. & $\boldsymbol{\bar{\mathcal{X}}}$ & The DFT of $\boldsymbol{\mathcal{X}}$.\\
\hline
$\mathbb{T}$ & A set. & $\boldsymbol{\mathcal{X}}(:,:,k)$ or $\boldsymbol{X}^{(k)}$ & The $k$-th frontal slice of $\boldsymbol{\mathcal{X}}$. & $\|\boldsymbol{\mathcal{X}}\|_{F}$ & $\sqrt{\sum_{ijk}|\boldsymbol{\mathcal{X}}_{ijk}|^{2}}$.\\

\hline
\hline
\end{tabular}
\label{notations}
\end{table*}

\begin{figure}[ht]
\begin{center}
{\includegraphics[width=0.7\textwidth]{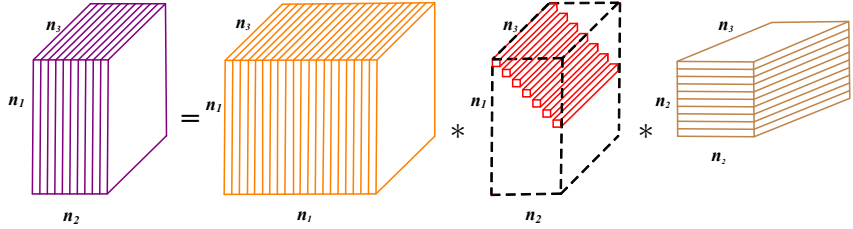}}
\end{center}
\vspace*{-14pt}
\caption{An illustration of the t-SVD of an $n_{1}\times n_{2}\times n_{3}$ tensor.}\label{illustration of the t-SVD}
\end{figure}

With the above notations, we first introduce three basic concepts of tensor algebra which will be used later.
\definition[t-SVD \cite{Kilmer2011Factorization}]\label{t-SVD}{Let $\boldsymbol{\mathcal{X}}\in\mathbb{R}^{n_{1}\times n_{2}\times n_{3}}$, the t-SVD factorization of tensor $\boldsymbol{\mathcal{X}}$ is
\begin{equation*}
\boldsymbol{\mathcal{X}}=\boldsymbol{\mathcal{U}} \ast \boldsymbol{\mathcal{S}} \ast \boldsymbol{\mathcal{V}}^{T},
\end{equation*}
where $\boldsymbol{\mathcal{U}}\in\mathbb{R}^{n_{1}\times n_{1}\times n_{3}}$ and $\boldsymbol{\mathcal{V}}\in\mathbb{R}^{n_{2}\times n_{2}\times n_{3}}$ are orthogonal, $\boldsymbol{\mathcal{S}}\in\mathbb{R}^{n_{1}\times n_{2}\times n_{3}}$ is an F-diagonal tensor. Figure \ref{illustration of the t-SVD} illustrates the t-SVD factorization.}

\definition[Tensor tubal rank \cite{kilmer2013third}]\label{Tensor tubal rank}{For $\boldsymbol{\mathcal{X}}\in\mathbb{R}^{n_{1}\times n_{2}\times n_{3}}$, the tensor tubal rank, denoted
as $\text{rank}_{t}(\boldsymbol{\mathcal{X}})$, is defined as the number of nonzero singular tubes of $\boldsymbol{\mathcal{S}}$, where $\boldsymbol{\mathcal{S}}$ is from the t-SVD of $\boldsymbol{\mathcal{X}}=\boldsymbol{\mathcal{U}} \ast \boldsymbol{\mathcal{S}} \ast \boldsymbol{\mathcal{V}}^{T}$. We can write
\begin{equation*}
\text{rank}_{t}(\boldsymbol{\mathcal{X}})=\sharp\left\{i:\boldsymbol{\mathcal{S}}(i,i,:)\neq\boldsymbol{0}\right\}.
\end{equation*}}

\definition[Tensor nuclear norm \cite{lu2018tensor}]\label{Tensor nuclear norm}{
Let $\boldsymbol{\mathcal{X}}=\boldsymbol{\mathcal{U}} \ast \boldsymbol{\mathcal{S}} \ast \boldsymbol{\mathcal{V}}^{T}$ be the t-SVD of $\boldsymbol{\mathcal{X}}\in\mathbb{R}^{n_{1}\times n_{2}\times n_{3}}$. The tensor nuclear norm of $\boldsymbol{\mathcal{X}}$ is defined as
\begin{equation*}
  \|\boldsymbol{\mathcal{X}}\|_{\circledast}:=\sum_{i=1}^{r}\boldsymbol{\mathcal{S}}(i,i,1),
\end{equation*}
where $r=\text{rank}_{t}(\boldsymbol{\mathcal{X}})$.}

\section{Probabilistic tools}
\label{Probabilistic tools}
This paper aims to answer which linear maps $\boldsymbol{\mathfrak{M}}$ satisfy the t-RIP. We will analyze this question from a more general perspective by considering the class of sub-Gaussian distributions. To this end, we first introduce some probabilistic tools that will be required for our results.
\definition[Sub-Gaussian random variables \cite{buldygin2000metric}]{A random variable $\xi$ is called sub-Gaussian if there exists a number $\alpha\in[0, \infty)$ such that the inequality}
\begin{equation*}
  \mathds{E}\exp(\theta\xi)\leq\exp\left(\frac{\alpha^{2}\theta^{2}}{2}\right)
\end{equation*}
holds for all $\theta\in\mathbb{R}$, and we denote that $\xi$ satisfies the above formula by $\xi\sim\rm Sub(\alpha^{2})$.
\remark{Sub-Gaussian distributions is a wider class of distributions as it contains zero-mean Gaussian distributions, symmetric Bernoulli distributions and all zero-mean bounded distributions. For example, if $\zeta$ is a Gaussian random variable with zero-mean and variance $\sigma^{2}$}, then $\zeta$ is also a sub-Gaussian random variable, i.e., $\zeta\sim\rm Sub(\sigma^{2})$. Therefore, we require that the distribution of all elements (tensors with size $n_{1} \times n_{2} \times n_{3} \times m$) of the measurement ensemble $\boldsymbol{\mathfrak{M}}: \mathbb{R}^{n_{1} \times n_{2} \times n_{3}}\rightarrow\mathbb{R}^{m}$ is a sub-Gaussian distribution.

Next, we provide some instrumental theoretical skills for the analysis of our main results which include $\varepsilon$-net, covering numbers, $\gamma_{\tau}$-functional and concentration inequalities.
\definition[$\varepsilon$-net \cite{Vershynin2018High}]{For a metric space $(\mathbb{T}, d)$, $\mathbb{W}\subset\mathbb{T}$ and $\tilde{\mathbb{W}}\subseteq\mathbb{W}$, if each element in $\mathbb{W}$ is within distance $\varepsilon$ ($\varepsilon>0$) of some elements of $\tilde{\mathbb{W}}$, i.e.
\begin{equation*}
  \forall t\in\mathbb{W},\;\exists t_{0}\in\tilde{\mathbb{W}}:\;d(t,t_{0}) \leq \varepsilon,
\end{equation*}
then the subset $\tilde{\mathbb{W}}$ is referred to as an $\varepsilon$-net of $\mathbb{W}$, denoted $\tilde{\mathbb{W}}_{\varepsilon}$.}

\remark{Throughout the article, we consider that $\mathbb{T}=\mathbb{R}^{n_{1} \times n_{2} \times n_{3}}$ and $d$ is the Euclidean distance, i.e. $d(\boldsymbol{\mathcal{X}},\boldsymbol{\mathcal{Y}})=\|\boldsymbol{\mathcal{X}}-\boldsymbol{\mathcal{Y}}\|,\;\boldsymbol{\mathcal{X}},\boldsymbol{\mathcal{Y}}\in\mathbb{R}^{n_{1} \times n_{2} \times n_{3}}$.}

\definition[Covering numbers \cite{Vershynin2018High}]{Let $\mathbb{W}$ be a subset of metric space $(\mathbb{T}, d)$. For $\varepsilon>0$, the covering number $\mathcal{N}(\mathbb{W}; d; \varepsilon)$ of $\mathbb{W}$ is defined as the smallest possible cardinality of an $\varepsilon$-net of $\mathbb{W}$.
}
\lemma[Covering numbers and volume \cite{Vershynin2018High}]\label{Covering numbers and volume}{If $\mathbb{W}$ be a subset of metric space $(\mathbb{R}^{n_{1} \times n_{2} \times n_{3}}, \|\cdot\|)$, then for $\varepsilon>0$, we have
\begin{equation*}
  \mathcal{N}(\mathbb{W}; d; \varepsilon)\leq\frac{\rm Vol(\mathbb{W}+\frac{\varepsilon}{2}\mathbb{K})}{\rm Vol(\frac{\varepsilon}{2}\mathbb{K})},
\end{equation*}
where $\rm Vol(\cdot)$ is the volume in $\mathbb{R}^{n_{1} \times n_{2} \times n_{3}}$ and $\frac{\varepsilon}{2}\mathbb{K}$ is Euclidean ball with radius $\varepsilon/2$.}

\remark{Note that when $\mathbb{W}$ is a unit Euclidean ball in $n_{1} \times n_{2} \times n_{3}$ dimensions (or it is the surface of the unit Euclidean ball), $\mathbb{W}+\frac{\varepsilon}{2}\mathbb{K}$ is contained in the $1+\varepsilon/2$ ball. If we assume that $\varepsilon\in(0,1]$, then we have the following crucial inequality,
\begin{equation}\label{3 dimensions Covering numbers}
  \mathcal{N}(\mathbb{W};d;\varepsilon)\leq\frac{(1+\varepsilon/2)^{n_{1}n_{2}n_{3}}}{(\varepsilon/2)^{n_{1}n_{2}n_{3}}}=\left(\frac{2+\varepsilon}{\varepsilon}\right)^{n_{1}n_{2}n_{3}}
  \leq(3/\varepsilon)^{n_{1}n_{2}n_{3}},
\end{equation}
which will is employed repetitively.}

It is useful to observe that the tensor restricted isometry constants $\delta_{r}$ can be expressed as a random variable $\xi$ as follows
\begin{equation*}
  \xi=\sup_{\boldsymbol{Z}\in\mathbb{Z}}\left|\|\boldsymbol{Z}\boldsymbol{\zeta}\|_{2}^{2}-\mathbb{E}\|\boldsymbol{Z}\boldsymbol{\zeta}\|_{2}^{2}\right|,
\end{equation*}
where $\mathbb{Z}$ is a set of matrices and $\boldsymbol{\zeta}$ is a sub-Gaussian vector (see the proof of Theorem \ref{main results1} for details). In order to obtain deviation bounds for random variables $\xi$ of this form in terms of a complexity parameter of the set of matrices $\mathbb{Z}$, we need to introduce the complexity parameter, i.e., Talagrand's $\gamma_{\tau}$-functional.
\definition[$\gamma_{\tau}$-functional \cite{talagrand2014Upper,Krahmer2014Suprema,Dirksen2015Tail}]\label{gamma functional}{Given a metric space $(\mathbb{T}, d)$, a collection of subsets of $\mathbb{T}$, $\{\mathbb{T}_{k}: k\geq0\}$, is referred to as an admissible sequence if $|\mathbb{T}_{0}|=1$ and $|\mathbb{T}_{k}|\leq2^{2^{k}}$ for every $k\geq1$, then the $\gamma_{\tau}$-functional with any $\tau\geq1$ of $\mathbb{T}$ is defined by
\begin{equation*}
  \gamma_{\tau}(\mathbb{T}, d)=\inf\sup_{t\in \mathbb{T}}\sum_{k=0}^{\infty}2^{k/\tau}d(t,\mathbb{T}_{k}),
\end{equation*}
where the infimum is taken in regard to all admissible sequences of $\mathbb{T}$ and $d(t,\mathbb{T}_{k})=\inf_{t_{0}\in\mathbb{T}_{k}}d(t,t_{0})$.}

In this paper, we mainly focus on the $\gamma_{2}$-functional of a set of matrices $\mathbb{Z}$ with the operator norm. The proof of our results requires the use of the covering number to give the bound of $\gamma_{2}$-functional.
In order to do this, we will utilize the Schatten spaces. Its detailed definition is as follows:
\begin{equation*}
  \|\boldsymbol{Z}\|_{S^{q}}=\left(\Tr\left(\boldsymbol{Z}^{T}\boldsymbol{Z}\right)^{q/2}\right)^{1/q},\quad(1\leq q<\infty)
\end{equation*}
and $\|\boldsymbol{Z}\|_{S^{\infty}}=\|\boldsymbol{Z}\|_{2\rightarrow2}$ are defined as the Schatten norms of a given matrix $\boldsymbol{Z}$, and
\begin{equation*}
  \bar{\Delta}_{q}(\mathbb{Z})=\sup_{\boldsymbol{Z}\in\mathbb{Z}}\|\boldsymbol{Z}\|_{S^{q}},\quad(1\leq q\leq\infty)
\end{equation*}
is defined as the radius of any set $\mathbb{Z}$ of matrices. Especially, $\bar{\Delta}_{2}(\mathbb{Z})=\sup_{\boldsymbol{Z}\in\mathbb{Z}}\|\boldsymbol{Z}\|_{S^{2}}=\sup_{\boldsymbol{Z}\in\mathbb{Z}}\|\boldsymbol{Z}\|_{F}:=\bar{\Delta}_{F}(\cdot)$.
With these notions, for a given metric space $(\mathds{Z}, \|\cdot\|_{2\rightarrow2})$ and $\nu>0$ with the covering number $\mathcal{N}(\mathds{Z}, \|\cdot\|_{2\rightarrow2}, \nu)$, by exploiting the Dudley type integral, we have the following inequality for $\gamma_{2}$-functional
\begin{equation}\label{Dudley type integral}
  \gamma_{2}(\mathbb{Z}, \|\cdot\|_{2\rightarrow2})\leq c\int_{0}^{\bar{\Delta}_{2\rightarrow2}(\mathbb{Z})}\sqrt{\log\mathcal{N}(\mathbb{Z}, \|\cdot\|_{2\rightarrow2}, \nu)}d\nu,
\end{equation}
where $c$ is a universal constant.

In CS, the following concentration inequality which involves $\gamma_{2}$-functional is often adopted to estimate the deviation bound of $\xi=\sup_{\boldsymbol{Z}\in\mathbb{Z}}\left|\|\boldsymbol{Z}\boldsymbol{\zeta}\|_{2}^{2}-\mathbb{E}\|\boldsymbol{Z}\boldsymbol{\zeta}\|_{2}^{2}\right|$. We will also make use of this important result.
\lemma[see \cite{rauhut2017low}]\label{Suprema of Chaos Processes}{Suppose that $\boldsymbol{\zeta}$ is a random vector whose entries $\zeta_{j}\overset{i.i.d.}{\sim}\rm Sub(\alpha^{2})$ with mean $0$ and variance $1$, where i.i.d. is the abbreviation of independent and identically distributed. Let $\mathbb{Z}$ be a set of matrices, and
\begin{align*}
  E=&\gamma_{2}(\mathbb{Z}, \|\cdot\|_{2\rightarrow2})\left(\gamma_{2}(\mathbb{Z}, \|\cdot\|_{2\rightarrow2})+\bar{\Delta}_{F}(\mathbb{Z})\right)+\bar{\Delta}_{F}(\mathbb{Z})\bar{\Delta}_{2\rightarrow2}(\mathbb{Z}),\\
  V=&\bar{\Delta}_{4}^{2}(\mathbb{Z}),\quad \rm and \quad U=\bar{\Delta}_{2\rightarrow2}^{2}(\mathbb{Z}).
\end{align*}
Then, there exist constants $c_{1}$, $c_{2}$ depending only on $\alpha$ such that for all $t>0$,
\begin{equation*}
  \mathds{P}\left(\sup_{\boldsymbol{Z}\in\mathbb{Z}}\left|\|\boldsymbol{Z}\boldsymbol{\zeta}\|_{2}^{2}-\mathbb{E}\|\boldsymbol{Z}\boldsymbol{\zeta}\|_{2}^{2}\right|\geq c_{1}E+t\right)\leq2\exp\left(-c_{2}\min\left\{\frac{t^{2}}{V^{2}},\frac{t}{U}\right\}\right).
\end{equation*}}

\section{Main results}
\label{Main results}
In this section, we will show that the t-RIP \eqref{t-RIP} holds with high probability for certain linear maps from a large class of random distributions satisfying the required number of measurements. We first compute the covering number of the set of tensors whose tubal rank is at most $r$ and Frobenius norm is $1$.
\lemma[Covering number for low-tubal-rank tensors]\label{Covering number for low-tubal-rank tensors}{For a set
\begin{equation*}
 \mathbb{W}_{r}=\left\{\boldsymbol{\mathcal{X}}\in\mathbb{R}^{n_{1}\times n_{2}\times n_{3}}: \rank\nolimits_{t}(\boldsymbol{\mathcal{X}})\leq r, \|\boldsymbol{\mathcal{X}}\|_{F}=1\right\},
\end{equation*}
there exists an $\varepsilon$-net $\tilde{\mathbb{W}}_{\varepsilon}\subset\mathbb{W}_{r}$ in regard to the Frobenius norm obeying
\begin{equation}\label{volumetric bound}
  \mathcal{N}(\mathbb{W}_{r}; \|\cdot\|_{F}; \varepsilon)\leq (9/\varepsilon)^{r(n_{1}+n_{2}+1)n_{3}}.
\end{equation}}
\remark{Lemma \ref{Covering number for low-tubal-rank tensors} leads to an important consequence of volumetric bound \eqref{volumetric bound} that is the covering numbers of the collection of low-tubal-rank tensors of interest and plays a key role in the proof of Theorem \ref{main results1}. Besides, note that the proof of Lemma \ref{Covering number for low-tubal-rank tensors} is based on the t-product and t-SVD whose definitions are consistent with matrix cases. Benefiting from the good property of the t-product and t-SVD, the bound \eqref{volumetric bound} can reduce to the corresponding result in low-rank matrix \cite{candes2011tight} when $n_{3}=1$.}

\begin{proof}
Here we take the proof strategy of Lemma 3.1 in \cite{candes2011tight} and modify it to accommodate our t-SVD.
For any $\boldsymbol{\mathcal{X}}\in\mathbb{W}_{r}$, we have the skinny t-SVD
\begin{equation*}
  \boldsymbol{\mathcal{X}}\overset{\rm t-SVD}{=}\boldsymbol{\mathcal{U}} \ast \boldsymbol{\mathcal{S}} \ast \boldsymbol{\mathcal{V}}^{T},
\end{equation*}
where $\boldsymbol{\mathcal{U}}\in\mathbb{R}^{n_{1}\times r\times n_{3}}$ and $\boldsymbol{\mathcal{V}}\in\mathbb{R}^{n_{2}\times r\times n_{3}}$ are two orthogonal tensors and $\boldsymbol{\mathcal{S}}\in\mathbb{R}^{r\times r\times n_{3}}$ is an F-diagonal tensor. Since
\begin{equation*}
  \|\boldsymbol{\mathcal{X}}\|_{F}=\frac{1}{\sqrt{n_{3}}}\|\boldsymbol{\bar{X}}\|_{F}=\frac{1}{\sqrt{n_{3}}}\|\boldsymbol{\bar{U}}\boldsymbol{\bar{S}}\boldsymbol{\bar{V}}^{T}\|_{F}=1,
\end{equation*}
so we have
\begin{equation*}
  \|\boldsymbol{\mathcal{S}}\|_{F}=\frac{1}{\sqrt{n_{3}}}\|\boldsymbol{\bar{S}}\|_{F}=\frac{1}{\sqrt{n_{3}}}\cdot\sqrt{n_{3}}=1.
\end{equation*}
We first construct $\varepsilon$-nets for sets of $\boldsymbol{\mathcal{U}}$, $\boldsymbol{\mathcal{V}}$ and $\boldsymbol{\mathcal{S}}$ respectively, and then achieve the purpose of covering $\mathbb{W}_{r}$. Without loss of generality, we may assume that $n_{1}=n_{2}=n$ since the adjustments for the general case will be obvious.

Let $\mathbb{F}=\{\boldsymbol{\mathcal{S}}\in\mathbb{R}^{r\times r\times n_{3}}: \|\boldsymbol{\mathcal{S}}\|_{F}=1\}$ be the set of F-diagonal tensors whose first frontal slice has nonnegative and nonincreasing diagonal entries. According to Lemma \ref{Covering numbers and volume} and \eqref{3 dimensions Covering numbers}, there exists an $\varepsilon/3$-net $\tilde{\mathbb{F}}_{\varepsilon/3}$
for $\mathbb{F}$ with $\mathcal{N}(\mathbb{F}; \|\cdot\|_{F}; \varepsilon/3)\leq(9/\varepsilon)^{n_{3}r}$. And then we let $\mathbb{G}=\{\boldsymbol{\mathcal{U}}\in\mathbb{R}^{n\times r\times n_{3}}: \boldsymbol{\mathcal{U}}^{T}\ast\boldsymbol{\mathcal{U}}=\boldsymbol{\mathcal{I}}\}$ and use the notation $\boldsymbol{\mathcal{U}}(:,j,:)$ to denote the $j$th lateral slice of $\boldsymbol{\mathcal{U}}$, i.e., a tensor in $\mathbb{R}^{n\times 1\times n_{3}}$. Definition 3.6 in \cite{kilmer2013third} shows that $\boldsymbol{\mathcal{U}}\in\mathbb{R}^{n\times r\times n_{3}}$ is an orthogonal tensor if and only if the lateral slices $\{\boldsymbol{\mathcal{U}}(:,1,:),\boldsymbol{\mathcal{U}}(:,2,:),\cdots,\boldsymbol{\mathcal{U}}(:,r,:)\}$
form an orthonormal set of matrices with $\|\boldsymbol{\mathcal{U}}(:,j,:)\|_{F}=1$. Therefore, it is less difficult to know that $\mathbb{G}$ is a subset of the unit ball under the following norm
\begin{equation*}
  \|\boldsymbol{\mathcal{U}}\|_{1,F}:=\max_{j}\|\boldsymbol{\mathcal{U}}(:,j,:)\|_{F}.
\end{equation*}
Hence, due to \eqref{3 dimensions Covering numbers}, there is an $\varepsilon/3$-net $\tilde{\mathbb{G}}_{\varepsilon/3}$ for $\mathbb{G}$ satisfying $\mathcal{N}(\mathbb{G}; \|\cdot\|_{1,F}; \varepsilon/3)\leq(9/\varepsilon)^{nn_{3}r}$. Then we can construct an $\varepsilon$-net
\begin{equation*}
  \tilde{\mathbb{K}}_{\varepsilon}=\{\tilde{\boldsymbol{\mathcal{U}}} \ast \tilde{\boldsymbol{\mathcal{S}}} \ast \tilde{\boldsymbol{\mathcal{V}}}^{T}:\;\tilde{\boldsymbol{\mathcal{U}}}, \tilde{\boldsymbol{\mathcal{V}}}\in\tilde{\mathbb{G}}_{\varepsilon/3}, \tilde{\boldsymbol{\mathcal{S}}}\in\tilde{\mathbb{F}}_{\varepsilon/3}\}
\end{equation*}
such that the covering number of the corresponding set $\mathbb{K}$ satisfies
\begin{equation*}
  \mathcal{N}(\mathbb{K}; \|\cdot\|_{F}; \varepsilon)\leq\mathcal{N}(\mathbb{F}; \|\cdot\|_{F}; \varepsilon/3)\cdot\mathcal{N}^{2}(\mathbb{G}; \|\cdot\|_{1,F}; \varepsilon/3)\leq(9/\varepsilon)^{r(2n+1)n_{3}}.
\end{equation*}
The rest of the work is to prove that $\tilde{\mathbb{K}}_{\varepsilon}$ is an $\varepsilon$-net for the set $\mathbb{W}_{r}$, i.e., $\tilde{\mathbb{K}}_{\varepsilon}=\tilde{\mathbb{W}}_{\varepsilon}$. In other words, we need to prove that for any $\boldsymbol{\mathcal{X}}\in\mathbb{W}_{r}$,  there exists $\tilde{\boldsymbol{\mathcal{X}}}\in\tilde{\mathbb{K}}_{\varepsilon}$ with $\|\boldsymbol{\mathcal{X}}-\tilde{\boldsymbol{\mathcal{X}}}\|_{F}\leq\varepsilon$.

Next, let $\tilde{\boldsymbol{\mathcal{X}}}\in\tilde{\mathbb{K}}_{\varepsilon}$ with
\begin{equation*}
  \tilde{\boldsymbol{\mathcal{X}}}\overset{\rm t-SVD}{=}\tilde{\boldsymbol{\mathcal{U}}} \ast \tilde{\boldsymbol{\mathcal{S}}} \ast \tilde{\boldsymbol{\mathcal{V}}}^{T},
\end{equation*}
where $\tilde{\boldsymbol{\mathcal{U}}}, \tilde{\boldsymbol{\mathcal{V}}}\in\tilde{\mathbb{G}}_{\varepsilon/3}, \tilde{\boldsymbol{\mathcal{S}}}\in\tilde{\mathbb{F}}_{\varepsilon/3}$ satisfying $\|\boldsymbol{\mathcal{U}}-\tilde{\boldsymbol{\mathcal{U}}}\|_{1,F}\leq\varepsilon/3$, $\|\boldsymbol{\mathcal{V}}-\tilde{\boldsymbol{\mathcal{V}}}\|_{1,F}\leq\varepsilon/3$, and $\|\boldsymbol{\mathcal{S}}-\tilde{\boldsymbol{\mathcal{S}}}\|_{F}\leq\varepsilon/3$, then we have
\begin{equation*}
  \|\boldsymbol{\mathcal{X}}-\tilde{\boldsymbol{\mathcal{X}}}\|_{F}=\|\boldsymbol{\mathcal{U}} \ast \boldsymbol{\mathcal{S}} \ast \boldsymbol{\mathcal{V}}^{T}-\tilde{\boldsymbol{\mathcal{U}}} \ast \tilde{\boldsymbol{\mathcal{S}}} \ast \tilde{\boldsymbol{\mathcal{V}}}^{T}\|_{F}
  \leq\|(\boldsymbol{\mathcal{U}}-\tilde{\boldsymbol{\mathcal{U}}}) \ast \boldsymbol{\mathcal{S}} \ast \boldsymbol{\mathcal{V}}^{T}\|_{F}+\|\tilde{\boldsymbol{\mathcal{U}}} \ast (\boldsymbol{\mathcal{S}}-\tilde{\boldsymbol{\mathcal{S}}}) \ast \boldsymbol{\mathcal{V}}^{T}\|_{F}
  +\|\tilde{\boldsymbol{\mathcal{U}}} \ast \tilde{\boldsymbol{\mathcal{S}}} \ast (\boldsymbol{\mathcal{V}}-\tilde{\boldsymbol{\mathcal{V}}})^{T}\|_{F},
\end{equation*}
where the first inequality uses the triangle inequality.  Since Frobenius norm has the property of being invariant under orthogonal multiplication and $\boldsymbol{\mathcal{U}}$, $\boldsymbol{\mathcal{V}}$ are two orthogonal tensors, we thus obtain
\begin{equation*}
  \|\tilde{\boldsymbol{\mathcal{U}}} \ast (\boldsymbol{\mathcal{S}}-\tilde{\boldsymbol{\mathcal{S}}}) \ast \boldsymbol{\mathcal{V}}^{T}\|_{F}=\|\boldsymbol{\mathcal{S}}-\tilde{\boldsymbol{\mathcal{S}}}\|_{F}\leq\varepsilon/3
\end{equation*}
and
\begin{equation*}
  \|(\boldsymbol{\mathcal{U}}-\tilde{\boldsymbol{\mathcal{U}}}) \ast \boldsymbol{\mathcal{S}} \ast \boldsymbol{\mathcal{V}}^{T}\|_{F}=\|(\boldsymbol{\mathcal{U}}-\tilde{\boldsymbol{\mathcal{U}}}) \ast \boldsymbol{\mathcal{S}}\|_{F}\leq\|\boldsymbol{\mathcal{U}}-\tilde{\boldsymbol{\mathcal{U}}}\|_{F}\|\boldsymbol{\mathcal{S}}\|_{F}
  \leq\|\boldsymbol{\mathcal{U}}-\tilde{\boldsymbol{\mathcal{U}}}\|_{1,F}\|\boldsymbol{\mathcal{S}}\|_{F}
  \leq\varepsilon/3,
\end{equation*}
So similarly, we would find that  $\|\tilde{\boldsymbol{\mathcal{U}}} \ast \tilde{\boldsymbol{\mathcal{S}}} \ast (\boldsymbol{\mathcal{V}}-\tilde{\boldsymbol{\mathcal{V}}})^{T}\|_{F}\leq\varepsilon/3$. Thus, we conclude that $\|\boldsymbol{\mathcal{X}}-\tilde{\boldsymbol{\mathcal{X}}}\|_{F}\leq\varepsilon$. This completes the proof.
\end{proof}

We are in the position to state our main results.
\theorem\label{main results1}{Fix $\delta, \varepsilon\in(0,1)$ and let $\boldsymbol{\mathcal{X}}\in\mathbb{R}^{n_{1} \times n_{2} \times n_{3}}$ be an any given third-order tensor whose tubal rank is at most $r$, then a random draw of a sub-Gaussian measurement ensemble $\boldsymbol{\mathfrak{M}}: \mathbb{R}^{n_{1} \times n_{2} \times n_{3}}\rightarrow \mathbb{R}^{m}$ satisfies $\delta_{r}\leq\delta$ with probability at least $1-\varepsilon$ provided that
\begin{equation*}
  m\geq C\delta^{-2}\max\left\{r(n_{1}+n_{2}+1)n_{3}, \log(\varepsilon^{-1})\right\},
\end{equation*}
where the constant $C>0$ only depends on the sub-Gaussian parameter.}

The following is a trivial corollary but an important special case of Theorem \ref{main results1}.
\corollary\label{corollary}{Let $\boldsymbol{\mathfrak{M}}: \mathbb{R}^{n_{1} \times n_{2} \times n_{3}}\rightarrow \mathbb{R}^{m}$ be a zero-mean Gaussian or symmetric Bernoulli measurement ensemble. Then there exists a universal constant $C>0$ such that the tensor restricted isometry constant of $\boldsymbol{\mathfrak{M}}$ satisfies $\delta_{r}\leq\delta$ with probability at least $1-\varepsilon$ provided that
\begin{equation*}
  m\geq C\delta^{-2}\max\left\{r(n_{1}+n_{2}+1)n_{3}, \log(\varepsilon^{-1})\right\}.
\end{equation*}
}

\remark{Theorem \ref{main results1} tells us that a random sub-Gaussian measurement ensemble obeys \eqref{t-RIP}. We know that sub-gaussian distributions belong to a larger class of random distributions, including zero-mean Gaussian distributions, symmetric Bernoulli distributions and all zero-mean bounded distributions}. Thus, in some sense, Theorem \ref{main results1} completely characterizes the behavior of numerous random measurement ensembles in term of the t-RIP.

\remark{For $\boldsymbol{\mathcal{X}}\in\mathbb{R}^{n_{1} \times n_{2} \times n_{3}}$, its degrees of freedom are the same as $\boldsymbol{\bar{\mathcal{X}}}$ since the DFT is the invertible. Suppose $\text{rank}_{t}(\boldsymbol{\mathcal{X}})=r$, then we have $\rank(\boldsymbol{\bar{X}}^{(i)})\leq r$, $i = 1,\cdots,n_{3}$. Then $\boldsymbol{\bar{X}}^{(i)}$ has at most $r(n_{1}+n_{2}-r)$ degrees of freedom, and thus $\boldsymbol{\mathcal{X}}$ has at most $r(n_{1}+n_{2}-r)n_{3}$ degrees of freedom. So the required number of measurements is very reasonable compared with the degrees of freedom. This reason is consistent with the viewpoint of Recht et al. on Theorem 4.2 in \cite{recht2010guaranteed}.

Now we further ascertain the strength of the bound. Observe that $O(r(n_{1} + n_{2} +1)n_{3})$ can be rewritten as $O(n_{(1)}n_{3}r)$ where $n_{(1)}=\max(n_{1},n_{2})$. So the bound is order optimal compared with the degrees of freedom of a tensor with tubal rank $r$. In addition, the set of rank-$r$ matrices contains all those matrices restricted to have nonzero entries only in the first $r$ rows. In this case, $\boldsymbol{\bar{X}}^{(i)}$ has at most $n_{(1)}r$ degrees of freedom, and thus $\boldsymbol{\mathcal{X}}$ has at most $n_{(1)}n_{3}r$ degrees of freedom. So, the bound implies that one only needs a constant number of measurements per degree of freedom of the underlying rank-$r$ tensor in order to obtain the t-RIP at rank $r$. The reasonability of the minimal possible number of linear measurements is further verified by experiments in Section \ref{Numerical experiments}. The above explanation coincides with the statement on Theorem 2.3 in \cite{candes2011tight} by Cand$\grave{e}$s et al.}

\remark{If $n_{3} = 1$, the third-order tensor $\boldsymbol{\mathcal{X}}$ will reduce to a two-order tensor, i.e., a matrix. Accordingly, the tensor tubal rank will reduce to the matrix rank, and t-RIP will reduce to the Definition 2.1 in \cite{candes2011tight}. Thus the required number of measurements for random sub-Gaussian measurement ensembles in Theorem \ref{main results1} includes the results of Theorem 2.3 in \cite{candes2011tight} for LRMR.}

\remark{It is worth mentioning that there exists a good study on the low-tubal-rank tensor recovery from Gaussian measurements by Lu et al. \cite{Lu2018Exact}. They based on the Gaussian width rather than RIP to show that the sampling number $O(r(n_{1}+n_{2}-r)n_{3})$ (or $O(rn_{(1)}n_{3})$ is sufficient and order optimal compared with the degrees of freedom of a tensor with tubal rank $r$. As mentioned in Introduction, the recovery results provided by Lu et al. are probabilistic: they are valid for a random measurement ensemble $\boldsymbol{\mathfrak{M}}$ and with high probability. We may wonder if there exists a deterministic condition which can guarantee that a given operator $\boldsymbol{\mathfrak{M}}$ can be used for tensor recovery. So Theorem \ref{main results1} and Corollary \ref{corollary} in this paper are just inspired by this problem.}

\remark{In CS and LRMR, Gaussian random matrix or Bernoulli random matrix is often used as a universal measurement matrix (ensemble) because they satisfy vector RIP \cite{Candes2005Decoding} with high probability. Accordingly, Corollary \ref{corollary} guarantees that the zero-mean Gaussian or symmetric Bernoulli measurement ensemble can also be used for LRTR.}

\begin{proof}
Given a tensor $\boldsymbol{\mathcal{X}}\in\mathbb{R}^{n_{1} \times n_{2} \times n_{3}}$ and a measurement ensemble $\boldsymbol{\mathfrak{M}}: \mathbb{R}^{n_{1} \times n_{2} \times n_{3}}\rightarrow \mathbb{R}^{m}$, then we can construct a matrix of size $m\times n_{1}n_{2}n_{3}m$ as follow
\begin{align*}
  \boldsymbol{D}_{\boldsymbol{\mathcal{X}}}=\frac{1}{\sqrt{m}}\left(
                                                              \begin{array}{cccc}
                                                                \boldsymbol{x}^{T} & \boldsymbol{0} & \cdots & \boldsymbol{0} \\
                                                                \boldsymbol{0} & \boldsymbol{x}^{T} & \cdots & \boldsymbol{0} \\
                                                                \vdots & \vdots & \ddots & \vdots \\
                                                                \boldsymbol{0} & \boldsymbol{0} & \cdots & \boldsymbol{x}^{T} \\
                                                              \end{array}
                                                            \right),
\end{align*}
with $\boldsymbol{x}$ being the vectorized version of the tensor $\boldsymbol{\mathcal{X}}$. and by utilizing an $n_{1}n_{2}n_{3}m$-dimensional random vector $\boldsymbol{\zeta}$ whose entries $\zeta_{j}\overset{i.i.d.}{\sim}\rm Sub(\alpha^{2})$ with mean 0 and variance 1 to obtain the measurements, that is
\begin{equation*}
  \boldsymbol{\mathfrak{M}}(\boldsymbol{\mathcal{X}})=\boldsymbol{D}_{\boldsymbol{\mathcal{X}}}\boldsymbol{\zeta}.
\end{equation*}
Setting $\mathbb{W}_{r}=\left\{\boldsymbol{\mathcal{X}}\in\mathbb{R}^{n_{1}\times n_{2}\times n_{3}}: \rank\nolimits_{t}(\boldsymbol{\mathcal{X}})\leq r, \|\boldsymbol{\mathcal{X}}\|_{F}=1\right\}$, by \eqref{t-RIP}, the tensor restricted isometry constant of $\boldsymbol{\mathfrak{M}}$ is
\begin{equation*}
  \delta_{r}=\sup_{\boldsymbol{\mathcal{X}}\in\mathbb{W}_{r}}\left|\|\boldsymbol{D}_{\boldsymbol{\mathcal{X}}}\boldsymbol{\zeta}\|_{2}^{2}-\|\boldsymbol{\mathcal{X}}\|_{F}^{2}\right|.
\end{equation*}
Further, it is not hard to check that
\begin{align*}
  \mathbb{E}\|\boldsymbol{D}_{\boldsymbol{\mathcal{X}}}\boldsymbol{\zeta}\|_{2}^{2}=&\sum\mathbb{E}\Bigg\{\bigg(\frac{1}{\sqrt{m}}x_{1}\zeta_{1}+\frac{1}{\sqrt{m}}x_{2}\zeta_{2}+\cdots+\frac{1}{\sqrt{m}}x_{n_{1}n_{2}n_{3}}\zeta_{n_{1}n_{2}n_{3}}\bigg)^{2}\\
  &+\bigg(\frac{1}{\sqrt{m}}x_{1}\zeta_{n_{1}n_{2}n_{3}+1}+\frac{1}{\sqrt{m}}x_{2}\zeta_{n_{1}n_{2}n_{3}+2}+\cdots+\frac{1}{\sqrt{m}}x_{n_{1}n_{2}n_{3}}\zeta_{2n_{1}n_{2}n_{3}}\bigg)^{2}\\
  &+\cdots+\bigg(\frac{1}{\sqrt{m}}x_{1}\zeta_{(m-1)n_{1}n_{2}n_{3}+1}+\frac{1}{\sqrt{m}}x_{2}\zeta_{(m-1)n_{1}n_{2}n_{3}+2}+\cdots+\frac{1}{\sqrt{m}}x_{n_{1}n_{2}n_{3}}\zeta_{mn_{1}n_{2}n_{3}}\bigg)^{2}\Bigg\}\\
=&\frac{1}{m}\cdot m\sum\left(x_{1}^{2}+x_{2}^{2}+\cdot+x_{n_{1}n_{2}n_{3}}^{2}\right)\\
=&\|\boldsymbol{x}\|_{2}^{2}=\|\boldsymbol{\mathcal{X}}\|_{F}^{2}
\end{align*}
and therefore we can write
\begin{equation*}
  \delta_{r}=\sup_{\boldsymbol{\mathcal{X}}\in\mathbb{W}_{r}}\left|\|\boldsymbol{D}_{\boldsymbol{\mathcal{X}}}\boldsymbol{\zeta}\|_{2}^{2}-\mathbb{E}\|\boldsymbol{D}_{\boldsymbol{\mathcal{X}}}\boldsymbol{\zeta}\|_{2}^{2}\right|.
\end{equation*}
In order to apply Lemma \ref{Suprema of Chaos Processes} to estimate the probabilistic bound for the above expressions, we define the set $\mathbb{Z}:=\{\boldsymbol{D}_{\boldsymbol{\mathcal{X}}}:\boldsymbol{\mathcal{X}}\in\mathbb{W}_{r}\}$ in Lemma \ref{Suprema of Chaos Processes}. Then we need to check that the radii $\bar{\Delta}_{F}(\mathbb{Z})$, $\bar{\Delta}_{2\rightarrow2}(\mathbb{Z})$, and $\bar{\Delta}_{4}(\mathbb{Z})$ of the set $\mathbb{Z}$ and the complexity parameter---Talagrand's functional $\gamma_{2}(\mathbb{Z}, \|\cdot\|_{2\rightarrow2})$. Clearly, $\bar{\Delta}_{F}(\mathbb{Z})=1$ is on account of  $\|\boldsymbol{D}_{\boldsymbol{\mathcal{X}}}\|_{F}=\|\boldsymbol{\mathcal{X}}\|_{F}=1$ for all $\boldsymbol{\mathcal{X}}\in\mathbb{W}_{r}$. In addition, based on this fact that the operator norm of a block-diagonal matrix is the maximum of the operator norms of the diagonal blocks and the operator norm of a vector is its $\ell_{2}$ norm, we see that
\begin{equation*}
  \|\boldsymbol{D}_{\boldsymbol{\mathcal{X}}}\|_{2\rightarrow2}=\frac{1}{\sqrt{m}}\|\boldsymbol{x}\|_{2}=\frac{1}{\sqrt{m}}\|\boldsymbol{\mathcal{X}}\|_{F}=\frac{1}{\sqrt{m}}.
\end{equation*}
Thus, we have $\bar{\Delta}_{2\rightarrow2}(\mathbb{Z})=\frac{1}{\sqrt{m}}$. And because of
\begin{align*}
  m\boldsymbol{D}_{\boldsymbol{\mathcal{X}}}\boldsymbol{D}^{T}_{\boldsymbol{\mathcal{X}}}=\left(
                                                              \begin{array}{cccc}
                                                                \|\boldsymbol{x}\|_{2}^{2} & \boldsymbol{0} & \cdots & \boldsymbol{0} \\
                                                                \boldsymbol{0} & \|\boldsymbol{x}\|_{2}^{2} & \cdots & \boldsymbol{0} \\
                                                                \vdots & \vdots & \ddots & \vdots \\
                                                                \boldsymbol{0} & \boldsymbol{0} & \cdots & \|\boldsymbol{x}\|_{2}^{2} \\
                                                              \end{array}
                                                            \right)=\boldsymbol{I}_{m},
\end{align*}
for all $\boldsymbol{\mathcal{X}}\in\mathbb{W}_{r}$, we obtain
\begin{equation*}
  \|\boldsymbol{D}_{\boldsymbol{\mathcal{X}}}\|_{S^{4}}^{4}=\Tr\left[(\boldsymbol{D}^{T}_{\boldsymbol{\mathcal{X}}}\boldsymbol{D}_{\boldsymbol{\mathcal{X}}})^{2}\right]=\Tr\left[(\boldsymbol{D}_{\boldsymbol{\mathcal{X}}}\boldsymbol{D}^{T}_{\boldsymbol{\mathcal{X}}})^{2}\right]
  =\Tr\left[\left(\frac{1}{m}\boldsymbol{I}_{m}\right)^{2}\right]=\Tr\left(\frac{1}{m^{2}}\boldsymbol{I}_{m}\right)=\frac{1}{m},
\end{equation*}
for all $\boldsymbol{D}_{\boldsymbol{\mathcal{X}}}\in\mathbb{Z}$. This implies that $\bar{\Delta}_{4}^{2}(\mathbb{Z})=\sup_{\boldsymbol{D}_{\boldsymbol{\mathcal{X}}}\in\mathbb{Z}}\|\boldsymbol{D}_{\boldsymbol{\mathcal{X}}}\|_{S^{4}}^{2}=\frac{1}{\sqrt{m}}$. Furthermore, by exploiting the Dudley type integral \eqref{Dudley type integral} and bound \eqref{volumetric bound} for $\mathcal{N}(\mathbb{W}_{r}, \|\cdot\|_{F}, \nu)$, we obtain the bound of the $\gamma_{2}$-functional
\begin{align}\label{bound of function}
  \gamma_{2}(\mathbb{Z}, \|\cdot\|_{2\rightarrow2})
  &\leq c\int_{0}^{1/\sqrt{m}}\sqrt{\log\mathcal{N}(\mathbb{W}_{r}, \|\cdot\|_{F}/\sqrt{m}, \nu)}d\nu\nonumber\\
  &= c\frac{1}{\sqrt{m}}\int_{0}^{1}\sqrt{\log\mathcal{N}(\mathbb{W}_{r}, \|\cdot\|_{F}, \nu)}d\nu\nonumber\\
  &\leq c\frac{1}{\sqrt{m}}\int_{0}^{1}\sqrt{r(n_{1}+n_{2}+1)n_{3}\log(9/\nu)}d\nu\nonumber\\
  &=c\sqrt{\frac{r(n_{1}+n_{2}+1)n_{3}}{m}}\int_{0}^{1}\sqrt{\log(9/\nu)}d\nu\nonumber\\
  &\leq c^{'}\sqrt{\frac{r(n_{1}+n_{2}+1)n_{3}}{m}}.
\end{align}
where $c^{'}$ is a universal constant. Let us now compute the constants $E$, $V$ and $U$ in Lemma \ref{Suprema of Chaos Processes}. This gives
\begin{align*}
  E=&\gamma_{2}(\mathbb{Z}, \|\cdot\|_{2\rightarrow2})\left(\gamma_{2}(\mathbb{Z}, \|\cdot\|_{2\rightarrow2})+\bar{\Delta}_{F}(\mathbb{Z})\right)+\bar{\Delta}_{F}(\mathbb{Z})\bar{\Delta}_{2\rightarrow2}(\mathbb{Z})\\
=&\gamma_{2}^{2}(\mathbb{Z}, \|\cdot\|_{2\rightarrow2})+\gamma_{2}(\mathbb{Z}, \|\cdot\|_{2\rightarrow2})+\frac{1}{\sqrt{m}},\\
  V=&\bar{\Delta}_{4}^{2}(\mathbb{Z})=\frac{1}{\sqrt{m}},\quad \rm and \quad U=\bar{\Delta}_{2\rightarrow2}^{2}(\mathbb{Z})=\frac{1}{m}.
\end{align*}
Substituting \eqref{bound of function} into $E$, we have
\begin{align*}
  c_{1}E\leq&c_{1}\left\{(c^{'})^{2}\frac{r(n_{1}+n_{2}+1)n_{3}}{m}+c^{'}\sqrt{\frac{r(n_{1}+n_{2}+1)n_{3}}{m}}+\frac{1}{\sqrt{m}}\right\}\\
\leq& c_{1}c_{3}\sqrt{\frac{r(n_{1}+n_{2}+1)n_{3}}{m}}\\
=& c_{4}\sqrt{\frac{r(n_{1}+n_{2}+1)n_{3}}{m}},
\end{align*}
where $c_{3}$, $c_{4}$ are universal constants and the second inequality holds as long as the constant $c_{3}$ is chosen appropriately. Thus, we conclude that if
\begin{equation}\label{bound part1}
  m\geq c_{5}\delta^{-2}r(n_{1}+n_{2}+1)n_{3}
\end{equation}
and $c_{5}$ is a universal constant, then $c_{1}E\leq\delta/2$. Let $t=\delta/2$, we get
\begin{equation*}
2\exp\left(-c_{2}\min\left\{\frac{t^{2}}{V^{2}},\frac{t}{U}\right\}\right)=2\exp\left(-c_{2}\frac{\delta^{2}}{4}m\right)\leq\varepsilon
\end{equation*}
provided that
\begin{equation}\label{bound part2}
  m\geq c_{6}\delta^{-2}\log(\varepsilon^{-1})
\end{equation}
where $c_{6}$ is a universal constant and is chosen appropriately. Combining \eqref{bound part1} with \eqref{bound part2}, and utilizing Lemma \ref{Suprema of Chaos Processes}, we can obtain the bound on $m$ presented in Theorem \ref{main results1} and complete the proof.
\end{proof}

%

\section{Numerical experiments}
\label{Numerical experiments}
In CS, it has been proved that it is NP-hard to verify vector RIP \cite{Candes2005Decoding} for a specific random matrix directly \cite{bandeira2013certifying}. Similarly, it seems very complex to check whether a given instance of a random measurement ensemble fails to obey t-RIP. In this section, we perform three sets of experiments to verify that $O(r(n_{1}+n_{2}-r)n_{3})$ samples are sufficient for robust recovery of a low-tubal-rank tensor, thus indirectly demonstrating our main results.

We perform $\boldsymbol{y}=\boldsymbol{M}\vec(\boldsymbol{\mathcal{X}})+\boldsymbol{w}$ to get the linear noise measurements instead of $\boldsymbol{y}=\boldsymbol{\mathfrak{M}}(\boldsymbol{\mathcal{X}})+\boldsymbol{w}$ where $\rm vec(\boldsymbol{\mathcal{X}})$ is a long vector obtained by stacking the columns of $\boldsymbol{\mathcal{X}}$. In all experiments, $\boldsymbol{w}\in\mathbb{R}^{m}$ is the Gaussian white noise with mean $0$ and variance $0.01^{2}$. Then the RTNNM model \eqref{tensor nuclear norm min unconstrained} can be reformulated as
\begin{equation}\label{reshape RTNNM}
  \min \limits_{{\boldsymbol{\mathcal{X}}}\in\mathbb{R}^{n_{1} \times n_{2} \times n_{3}}}~\|\boldsymbol{\mathcal{X}}\|_{\circledast}+\frac{1}{2\lambda}\|\boldsymbol{y}-\boldsymbol{M}\rm vec(\boldsymbol{\mathcal{X}})\|_{2}^{2}.
\end{equation}
We adopt effective Algorithm 1 in \cite{Zhang2019RIP} to solve \eqref{reshape RTNNM}. We deem that the tensor $\hat{\boldsymbol{\mathcal{X}}}$ can be as a successful reconstruction for the original tensor $\boldsymbol{\mathcal{X}}$ from the measurements $\boldsymbol{y}$ if the relative error (abbreviation: RelError) satisfies $\|\hat{\boldsymbol{\mathcal{X}}}-\boldsymbol{\mathcal{X}}\|_{F}/\|\boldsymbol{\mathcal{X}}\|_{F}<10^{-3}$.

In the first experiment, we present numerical results for recovery of third-order tensors $\boldsymbol{\mathcal{X}}\in\mathbb{R}^{n_{1}\times n_{2}\times n_{3}}$ with different problem setups, i.e., different tensor sizes $n_{1}\times n_{2}\times n_{3}$, tubal ranks $r$, measurement ensembles $\boldsymbol{\mathfrak{M}}$ and sampling rate $m/(n_{1}n_{2}n_{3})$. We consider two sizes of $\boldsymbol{\mathcal{X}}\in\mathbb{R}^{n_{1}\times n_{2} \times n_{3}}$ and different tubal ranks: (a) $n_{1}=n_{2}=10$, $n_{3}=5$, $r_{1}=1$, $r_{2}=2$, $r_{3}=3$; (b) $n_{1}=n_{2}=20$, $n_{3}=5$, $r_{1}=2$, $r_{2}=4$, $r_{3}=6$. $\boldsymbol{M}\in\mathbb{R}^{m\times (n_{1}n_{2}n_{3})}$ is a measurement matrix with i.i.d. zero-mean Gaussian entries having variance $1/m$ or i.i.d. Bernoulli entries, i.e., $\mathds{P}(\boldsymbol{M}_{i,j}=\pm1/\sqrt{m})=1/2$.

\begin{figure*}[!htb]
\centering
\subfloat[]{\includegraphics[width=0.52\textwidth]{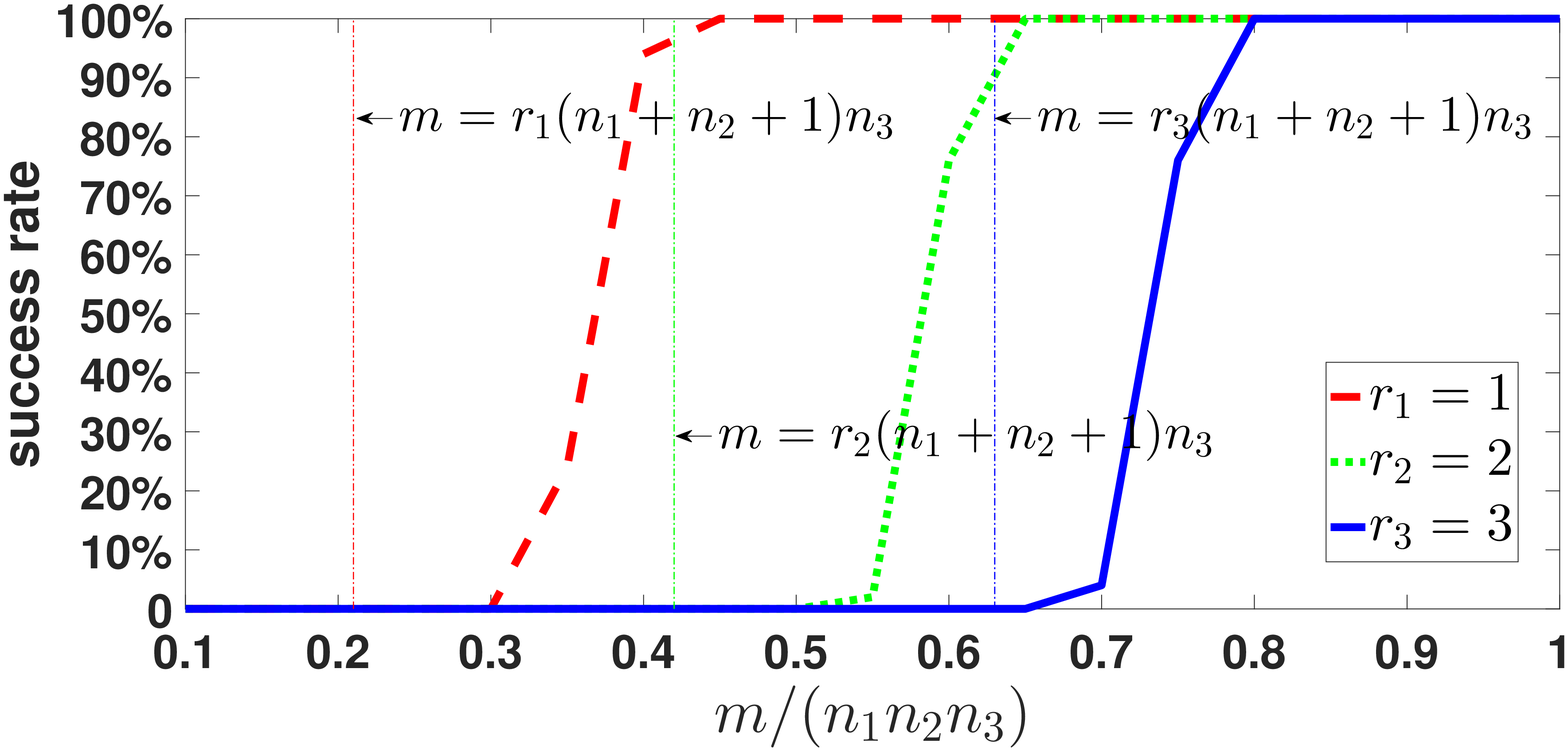}}
\subfloat[]{\includegraphics[width=0.52\textwidth]{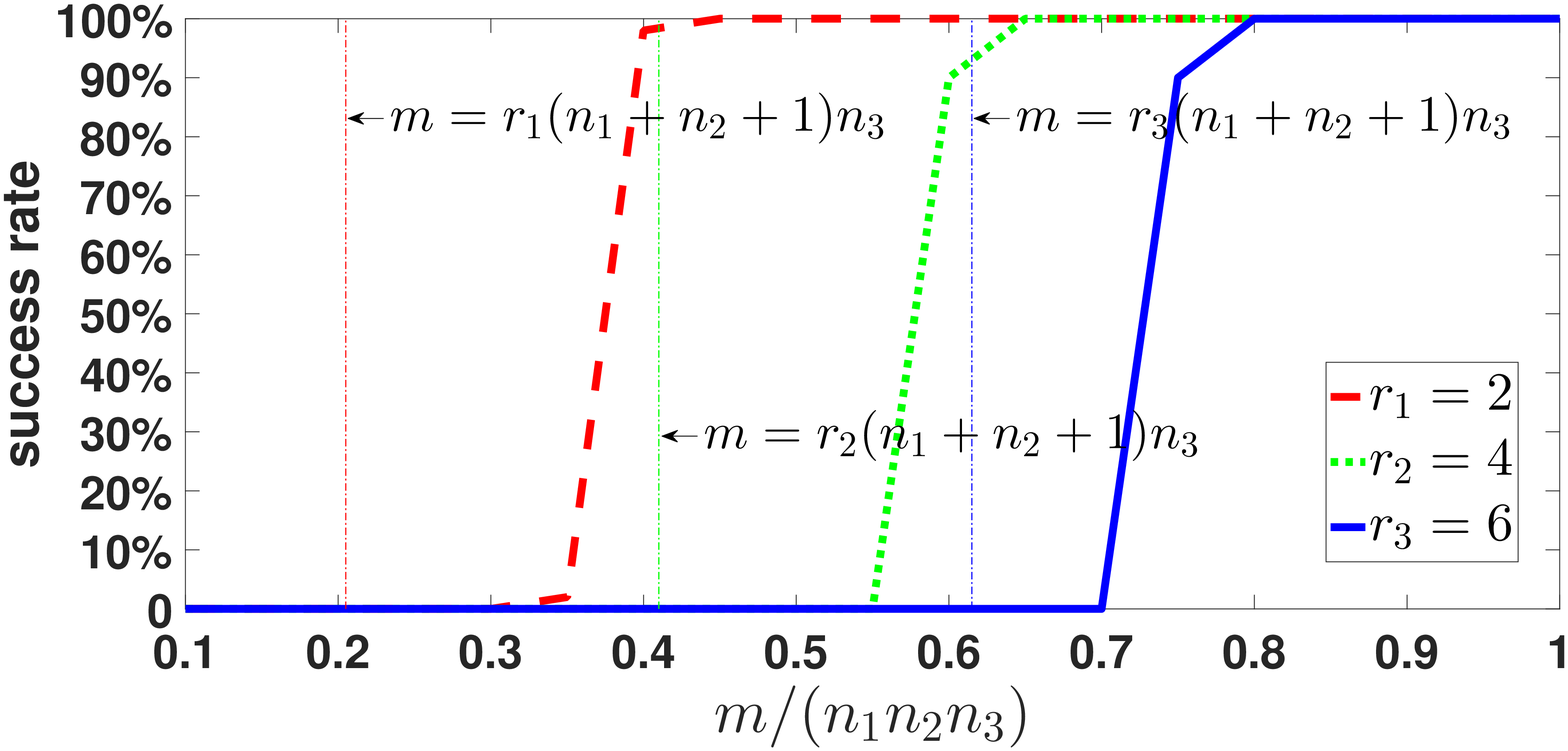}}
\vspace*{-10pt}
\caption{Successful recovery rate in 50 trials versus the sampling rate $m/(n_{1}n_{2}n_{3})$ for the random Gaussian measurements ensemble. There exist two sizes of $\boldsymbol{\mathcal{X}}\in\mathbb{R}^{n_{1}\times n_{2} \times n_{3}}$ with different tubal ranks: (a) $n_{1}=n_{2}=10$, $n_{3}=5$, $r_{1}=1$, $r_{2}=2$, $r_{3}=3$; (b) $n_{1}=n_{2}=20$, $n_{3}=5$, $r_{1}=2$, $r_{2}=4$, $r_{3}=6$. The minimum required sampling rate for successful recovery is indicated by the vertical lines.}\label{Gaussian measurements}
\end{figure*}

\begin{figure*}[!htb]
\centering
\subfloat[]{\includegraphics[width=0.52\textwidth]{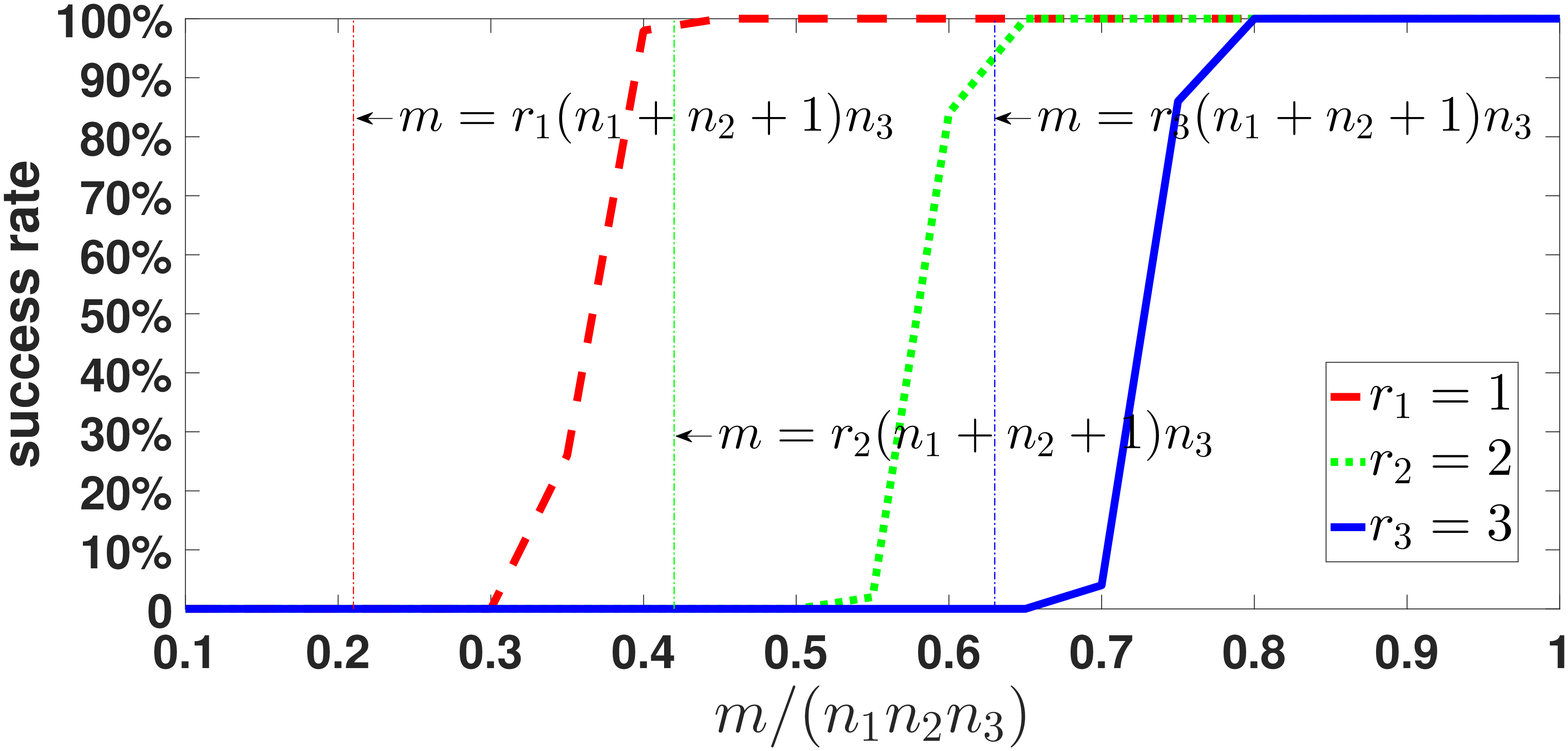}}
\subfloat[]{\includegraphics[width=0.52\textwidth]{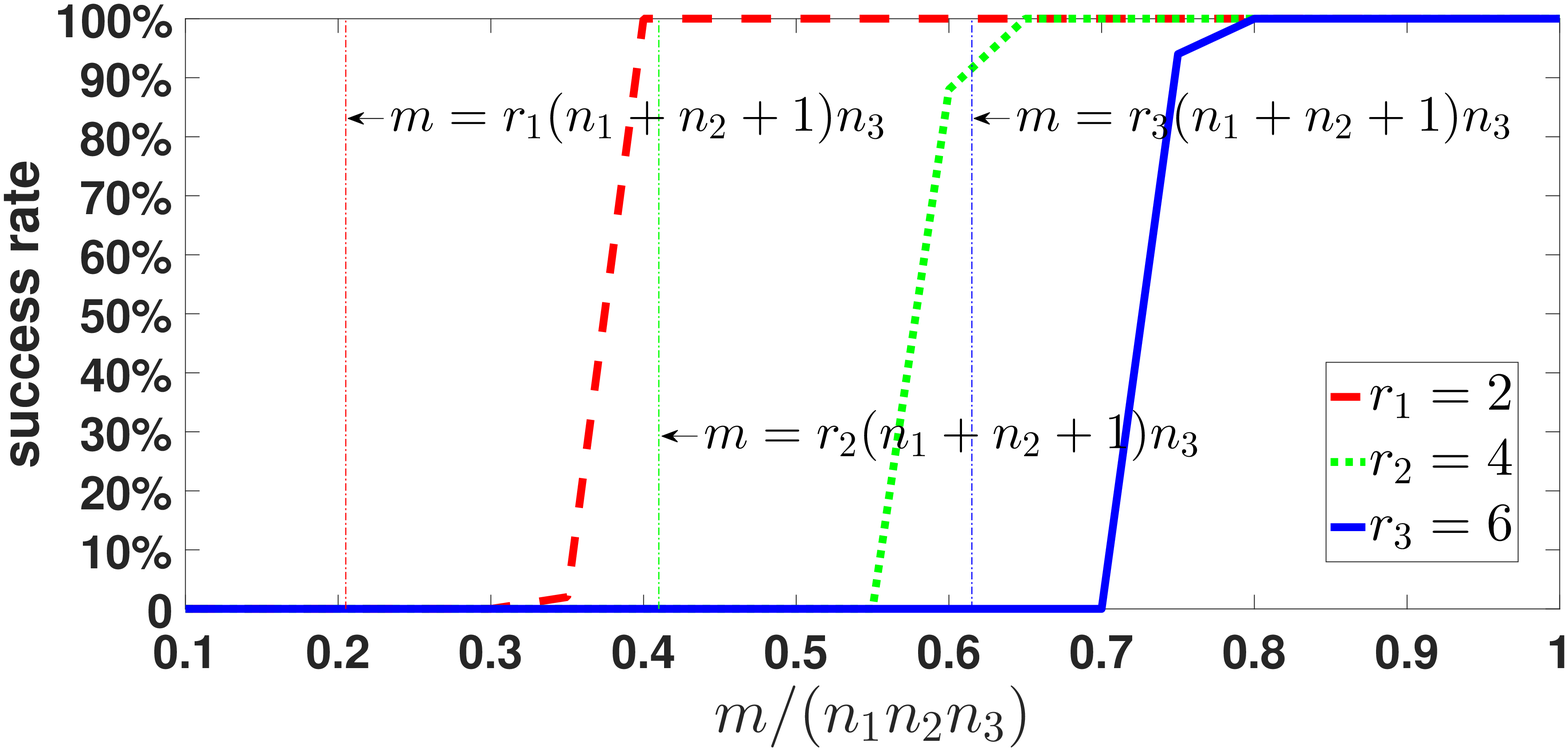}}
\vspace*{-10pt}
\caption{Successful recovery rate in 50 trials versus the sampling rate $m/(n_{1}n_{2}n_{3})$ for the random Bernoulli measurements ensemble. There exist two sizes of $\boldsymbol{\mathcal{X}}\in\mathbb{R}^{n_{1}\times n_{2} \times n_{3}}$ with different tubal ranks: (a) $n_{1}=n_{2}=10$, $n_{3}=5$, $r_{1}=1$, $r_{2}=2$, $r_{3}=3$; (b) $n_{1}=n_{2}=20$, $n_{3}=5$, $r_{1}=2$, $r_{2}=4$, $r_{3}=6$. The minimum required sampling rate for successful recovery is indicated by the vertical lines.}\label{Bernoulli measurements}
\end{figure*}

Figure \ref{Gaussian measurements} and Figure \ref{Bernoulli measurements} show the success rate of recovery in $50$ trials versus the sampling rate $m/(n_{1}n_{2}n_{3})$ for the random Gaussian measurements ensemble and random Bernoulli measurements ensemble, respectively. The minimum required sampling rate by theory (the minimum required number of measurements, i.e., $m=r(n_{1}+n_{2}+1)n_{3}$) for successful recovery is indicated by the vertical lines. All of the cases consistently show that the unknown tensor of size $n_{1} \times n_{2} \times n_{3}$ with tubal rank $r$ can be successfully recovered by solving \eqref{tensor nuclear norm min unconstrained} when the given number of measurements $m=\Omega(r(n_{1}+n_{2}+1)n_{3})$. This conclusion, combined with Theorem 4.1 in \cite{Zhang2019RIP}, verifies our Theorem 1. However, from Figure \ref{Gaussian measurements} and Figure \ref{Bernoulli measurements}, it is not difficult to find that there is a small gap between the required number of measurements by theory and that required by the experiment. This gap is allowed because there are many factors in the experiment such as the choice of algorithm, parameter setting, etc., which may cause this gap.

In the second experiment, we recover random tensors from Gaussian measurements of the size suggested by Theorem \ref{main results1}. Major experimental settings remain unchanged. Specifically, we let $n_{1}=n_{2}=n=10,20,30$, $n_{3}=5$ and $r=0.1n,0.2n,0.3n$. The number of measurements is set to $m=\rho r(n_{1}+n_{2}+1)n_{3}$ as in Theorem \ref{main results1} where $\rho=1,1.5,2$. The numerical results are presented in Table \ref{table 2}. It can be seen that as $\rho$ increases, that is, the number of measurements increases by a factor of that $r(n_{1}+n_{2}+1)n_{3}$, the relative error will become very small. Such a result indirectly verifies our theoretical result provided by Theorem \ref{main results1}.
\begin{table*}[!htb]
  \centering
  \fontsize{8}{8}\selectfont
  \begin{threeparttable}
  \caption{Robust low-tubal-rank tensor recovery from Gaussian measurements.}\label{table 2}
    \begin{tabular}{ccccccc}
    \multicolumn{7}{c}{$r=0.1n,m=\rho r(n_{1}+n_{2}+1)n_{3}$}\cr
    \toprule
    \multirow{2}{*}{$n$}&
    \multicolumn{2}{c}{$\rho=1$}&\multicolumn{2}{c}{$\rho=1.5$}&\multicolumn{2}{c}{$\rho=2$}\cr
    \cmidrule(lr){2-3} \cmidrule(lr){4-5} \cmidrule(lr){6-7}
    &$m$&RelError&$m$&RelError&$m$&RelError\cr
    \midrule
    $10$&        $105$&$0.6729$&$158$&$0.1061$&$210$&$3.7967e$$-5$\cr
    $20$&        $410$&$0.5884$&$615$&$0.2329$&$820$&$4.8508e$$-5$\cr
    $30$&        $915$&$0.5891$&$1373$&$0.2043$&$1830$&$4.3765e$$-5$\cr
    \bottomrule
    \multicolumn{7}{c}{$r=0.2n,m=\rho r(n_{1}+n_{2}+1)n_{3}$}\cr
    \toprule
    \multirow{2}{*}{$n$}&
    \multicolumn{2}{c}{$\rho=1$}&\multicolumn{2}{c}{$\rho=1.5$}&\multicolumn{2}{c}{$\rho=2$}\cr
    \cmidrule(lr){2-3} \cmidrule(lr){4-5} \cmidrule(lr){6-7}
    &$m$&RelError&$m$&RelError&$m$&RelError\cr
    \midrule
    $10$&        $210$&$0.4138$&$315$&$5.3116e-5$&$420$&$4.3057e$$-5$\cr
    $20$&        $820$&$0.4316$&$1230$&$6.2231e-5$&$1640$&$4.3732e$$-5$\cr
    $30$&        $1830$&$0.4305$&$2745$&$5.3885e-5$&$3660$&$3.2673e$$-5$\cr
    \bottomrule
    \multicolumn{7}{c}{$r=0.3n,m=\rho r(n_{1}+n_{2}+1)n_{3}$}\cr
    \toprule
    \multirow{2}{*}{$n$}&
    \multicolumn{2}{c}{$\rho=1$}&\multicolumn{2}{c}{$\rho=1.5$}&\multicolumn{2}{c}{$\rho=2$}\cr
    \cmidrule(lr){2-3} \cmidrule(lr){4-5} \cmidrule(lr){6-7}
    &$m$&RelError&$m$&RelError&$m$&RelError\cr
    \midrule
    $10$&        $315$&$0.2524$&$473$&$4.9014e-5$&$630$&$3.6912e$$-5$\cr
    $20$&        $1230$&$0.1846$&$1845$&$3.6913e-5$&$2460$&$2.7512e$$-5$\cr
    $30$&        $2745$&$0.2571$&$4118$&$3.2331e-5$&$5490$&$2.4352e$$-5$\cr
    \bottomrule
    \end{tabular}
    \end{threeparttable}
\end{table*}

\begin{figure*}[ht]
\begin{center}
{\includegraphics[width=1\textwidth]{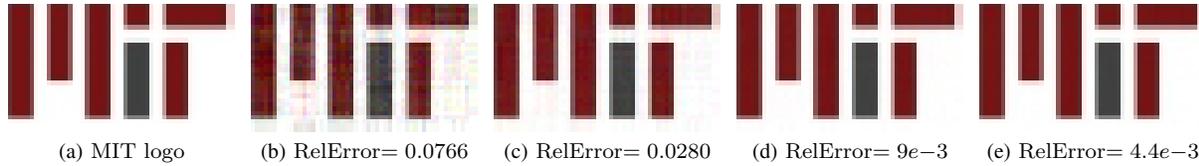}}
\end{center}
\vspace*{-14pt}
\caption{Example of recovered images using the Gaussian ensemble. (a) MIT logo test image. (b) 1260 measurements. (c) 1638 measurements. (d) 1890 measurements. (e) 2520 measurements. The total number of pixels is $30 \times 53 \times 3= 4770$.}\label{MIT}
\end{figure*}

To further illustrate the scaling of robust low-tubal-rank tensor recovery from Gaussian measurements, in the third experiment, we test on the MIT logo presented in Figure \ref{MIT} (a). It is a real third-order tensor data with a size of $30\times51\times3$ and a tubal rank of 5. The total number of pixels is $30 \times 53 \times 3= 4770$. Figure \ref{MIT} shows the reconstructed images from different Gaussian measurements. It is easy to find that when $m=1890$ and $m=2520$, which are equal to $1.5r(n_{1}+n_{2}+1)n_{3}$ and $2r(n_{1}+n_{2}+1)n_{3}$ respectively, the MIT logo is recovered perfectly with very small relative error. This experiment is in agreement with the result of Theorem \ref{main results1}.

\section{Conclusion and future work}
\label{Conclusion}
In this paper, using probabilistic arguments, we derive a reasonable lower bound on the number of measurements that makes the random sub-Gaussian measurement ensembles with high probability satisfy the t-RIP, defined by Zhang et al. \cite{Zhang2019RIP} in LRTR. Because sub-gaussian distributions belong to a larger class of random distributions, including zero-mean Gaussian distributions, symmetric Bernoulli distributions and all zero-mean bounded distributions, so the required number of measurements holds for Gaussian and Bernoulli measurement ensembles commonly used in experiments. These provide a theoretical basis for designing algorithms to solve \eqref{rank min} by TNNM \eqref{tensor nuclear norm min} and RTNNM\eqref{tensor nuclear norm min unconstrained}.

In future work, we will study the following tensor Schatten-$q$ nuclear norm minimization model and regularized tensor Schatten-$q$ nuclear norm minimization model $(0<q\leq1)$
\begin{equation*}
  \min \limits_{{\boldsymbol{\mathcal{X}}}\in\mathbb{R}^{n_{1} \times n_{2} \times n_{3}}}~\|\boldsymbol{\mathcal{X}}\|_{S_{q}}^{q},~~s.t.~~\|\boldsymbol{y}-\boldsymbol{\mathfrak{M}}(\boldsymbol{\mathcal{X}})\|_{2}\leq\epsilon,
\end{equation*}
and
\begin{equation*}
  \min \limits_{{\boldsymbol{\mathcal{X}}}\in\mathbb{R}^{n_{1} \times n_{2} \times n_{3}}}~\|\boldsymbol{\mathcal{X}}\|_{S_{q}}^{q}+\frac{1}{2\lambda}\|\boldsymbol{y}-\boldsymbol{\mathfrak{M}}(\boldsymbol{\mathcal{X}})\|_{2}^{2},
\end{equation*}
where $\|\boldsymbol{\mathcal{X}}\|_{S_{q}}^{q}=\frac{1}{n_{3}}\sum_{i=1}^{n_{3}}\|\boldsymbol{\bar{X}}^{(i)}\|_{S_{q}}^{q}$ is defined as the tensor Schatten-$q$ norm. In addition, we will extend the notion of the t-RIP to the tensor Schatten-$q$ Restricted Isometry Property and obtain the corresponding theoretical results.


\Acknowledgements{This work was supported by National Natural Science Foundation of China (Grant Nos. 61673015, 61273020), Fundamental Research Funds for the Central Universities (Grant Nos. XDJK2018C076, SWU1809002), China Postdoctoral Science Foundation (Grant No. 2018M643390) and Graduate Student Scientific Research Innovation Projects in Chongqing (Grant No. CYB19083).}

\end{document}